%% file: 1321.tex
\documentclass[runningheads]{llncs}
\usepackage{graphicx}
\usepackage{comment}
\usepackage{amsmath,amssymb} % define this before the line numbering.
\usepackage{color}

\usepackage{graphicx}
\usepackage{amsmath}
\usepackage{amssymb}

\usepackage{enumitem}
\usepackage{placeins}

\usepackage{url}
\usepackage{hyperref}
\usepackage{booktabs}

\usepackage{subfig}

\input{definitions}

\begin{document}
% \renewcommand\thelinenumber{\color[rgb]{0.2,0.5,0.8}\normalfont\sffamily\scriptsize\arabic{linenumber}\color[rgb]{0,0,0}}
% \renewcommand\makeLineNumber {\hss\thelinenumber\ \hspace{6mm} \rlap{\hskip\textwidth\ \hspace{6.5mm}\thelinenumber}}
% \linenumbers
\pagestyle{headings}
\mainmatter
\def\ECCVSubNumber{1321}  % Insert your submission number here

\title{From Shadow Segmentation to Shadow Removal} % Replace with your title

% INITIAL SUBMISSION 
%\begin{comment}
%\titlerunning{ECCV-20 submission ID \ECCVSubNumber} 
%\authorrunning{ECCV-20 submission ID \ECCVSubNumber} 
%\author{Anonymous ECCV submission}
%\institute{Paper ID \ECCVSubNumber}
%\end{comment}
%******************

% CAMERA READY SUBMISSION
%\begin{comment}
\titlerunning{From Shadow Segmentation to Shadow Removal}
% If the paper title is too long for the running head, you can set
% an abbreviated paper title here
%
\author{Hieu Le \and
Dimitris Samaras }
\authorrunning{H. Le  \& D. Samaras }
% First names are abbreviated in the running head.
% If there are more than two authors, 'et al.' is used.
%
\institute{Stony Brook University, Stony Brook, NY 11794, USA}
%\\
%\email{\{abc,lncs\}@uni-heidelberg.de}}
%\end{comment}
%******************
\maketitle

\begin{abstract}
The requirement for paired shadow and shadow-free images limits the size and diversity of shadow removal datasets and hinders the possibility of training large-scale, robust shadow removal algorithms. We propose a shadow removal method that can be trained using only shadow and non-shadow patches cropped from the shadow images themselves. Our method is trained via an adversarial framework, following a physical model of shadow formation. Our central contribution is a set of physics-based constraints that enables this adversarial training. Our method achieves competitive shadow removal results compared to state-of-the-art methods that are trained with fully paired shadow and shadow-free images. The advantages of our training regime are even more pronounced in shadow removal for videos. Our method can be fine-tuned on a testing video with only the shadow masks generated by a pre-trained shadow detector and outperforms state-of-the-art methods on this challenging test. We illustrate the advantages of our method on our proposed video shadow removal dataset. 
\keywords{Shadow Removal, GAN, Weakly-supervised, Illumination model, Unpaired, Image-to-Image.}
\end{abstract}

\section{Introduction}

%\mtodo{shadow editing, photo realistic vs shadow suppression for recognition}
Shadows are present in most natural images. Shadow effects make objects harder to detect or segment \cite{m_Le-etal-ECCV18}, and scenes with shadows are harder to process and analyze \cite{Le_2019_CVPR_Workshops}. Realistic shadow removal is an integral part of image editing \cite{Chuang2003} and can greatly improve performance on various computer vision tasks  \cite{Mller2019BrightnessCA,Su2016ShadowDA,Zhang2019ImprovingSS,LeICCV2017,Le2016GeodesicDH}, getting increased attention in recent years \cite{Qu_2017_CVPR,guo11,Gong16}.  Data-driven approaches using deep learning models have achieved remarkable performance on shadow removal \cite{Ding2019ARGANAR,Le-etal-ICCV19,Hu_2018_CVPR,hu_iccv2019mask,Wang_2018_CVPR,Zhang:AAA2020} thanks to  recent large-scale datasets \cite{Vicente-etal-ECCV16,Wang_2018_CVPR}. %\mtodo{shadow removal why}

\begin{figure}[t!]
	\centering
  \includegraphics[width=0.9\textwidth]{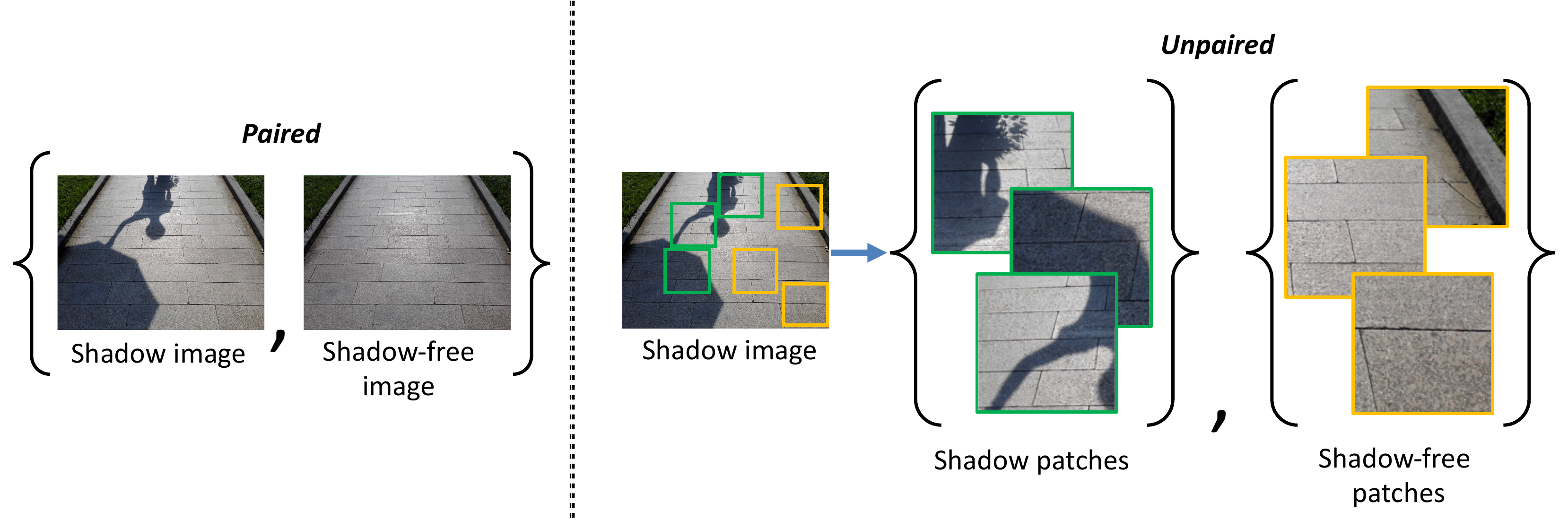}

  \caption{Paired training data (left) consists of training examples \{shadow, shadow-free\} images which are expensive to collect, lack diversity, and are sensitive to errors due to possible color mismatches between the two images. Note the slightly different color tone between the two images. In this paper, we propose to learn shadow removal from unpaired shadow  and non-shadow patches cropped from the same shadow image (right). This eliminates the need for shadow free images. }
  \label{fig:Teaser}
\end{figure}

Most of the current deep-learning  shadow removal approaches are  end-to-end mapping functions trained in a fully supervised manner. Such systems require pairs of shadow images and their shadow-free counter-parts as training signals. However, this type of data is cumbersome to obtain, lacks diversity, and is error-prone: all current shadow removal datasets exhibit color mismatches between the shadow images and their shadow-free ground truth (see Fig. \ref{fig:Teaser} - left panel). Moreover, there are no images with self-cast shadows because the occluders are never visible in the image in the  current data acquisition setups \cite{Wang_2018_CVPR,Qu_2017_CVPR,hu_iccv2019mask}.
This dependency on paired data significantly hinders  building  large-scale,  robust shadow-removal systems. A recent method trying to overcome this issue is  MaskShadow-GAN \cite{hu_iccv2019mask}, which learns shadow removal from unpaired shadow and shadow-free images. However, such  cycle-GAN \cite{CycleGAN2017} based systems usually require enough statistical similarity between the two sets of images \cite{Li2019AsymmetricGF,Choi2017StarGANUG}. %in order to enable the network to learn shadow removal rather than manipulating other visual parts of the image. % to fool the discriminator. 
This requirement can be hard to satisfy when capturing shadow-free images is tricky, such as shadow-free images of urban areas \cite{Shadow_Urban} or moving objects \cite{Shadow_tracking,Prati2003DetectingMS}.
%This significantly relaxs the data-dependent issue.
%MaskShadow-GAN learns a transformation between shadow and shadow-free images .   %
%\mtodo{discussion - Maskshadow -mismatch} In an effort to get a better statistic of the shadow image, we want to capture the information directly from the image itself.
%\mtodo{}

In this paper, we propose an alternative solution to the data dependency issue. We first observe that image patches alongside the shadow boundary contain critical information for shadow removal, including non-shadow, umbra and penumbra areas. They sufficiently reflect the characteristics of the shadowing effects, including the color differences between shadow and non-shadow areas as well as the gradual changes of the shadow effects across the shadow boundary \cite{Panagopoulospami13,panagopoulos2010estimating,guoPami}. If we further assume that the shadow effects are fairly consistent in the umbra areas, a patch-based shadow removal can be used to remove shadows in the whole image. Based on this observation, we propose training a patch-based shadow removal system for which we use unpaired shadow and non-shadow patches directly cropped from the shadow images themselves as training data. This approach eliminates the dependency on paired training data and opens up the possibility of handling different types of shadows, since it can be trained with any kind of shadow image. Compared to MaskShadow-GAN, shadow and non-shadow patches cropped from the same image naturally ensure significant statistical similarity. The only supervision required in this data processing scheme are the shadow masks, which are relatively easy to obtain, either manually, semi-interactively \cite{Vicente-etal-ECCV16,Gong16}, or automatically using shadow detection methods \cite{Ding2019ARGANAR,zhu18b,Zheng2019DistractionAwareSD,m_Le-etal-ECCV18}.
Automatic shadow detection is improving, with the main challenge being generalization across datasets. At some point, one can expect to get very good shadow masks automatically, which would allows training our shadow removal method with very little annotation cost.

In particular, to obtain shadow and shadow-free patches, we crop the shadow images into small overlapping patches of size $n\times n$ with a step size of $m$. Based on the shadow masks, we group these patches into three sets: a non-shadow set ($\mN$) containing patches having no shadow pixels, a shadow-boundary set ($\mB$) containing patches lying on the shadow boundaries, and a full-shadow set ($\mF$) containing patches where all pixels are in shadow.
With small enough patch size $n$ and  step size $m$, we can obtain enough training patches in each set. With this training set, we train a shadow removal system to learn a mapping  from patches in the shadow-boundary set $\mB$ to patches in the non-shadow set $\mN$. %Patches lying on the shadow boundary expose the intensity difference between shadow and non-shadow pixels. Here we assume that this color difference is somewhat consistent for the whole shadow image and thus we can remove the shadow from the whole image just by analyzing small patches.
%\mtodo{check inpainitng, image restoration papers}
%Here it is critical to notice that a shadow-boundary patch contains sufficient clues for shadow removal, including the color shift between shadow and non-shadow pixels as well as the gradual changes of the shadow effects across the shadow boundaries.
Essentially, this mapping needs to infer the color difference alongside the shadow edges, including the chromatic attributes of the light source and the smooth change of the shadow effects across the shadow boundary, in order to transform a shadow patch to a non-shadow patch.
This is, in spirit, similar to early shadow removal approaches that focus on shadow edges to remove shadows \cite{shadow_edge,Finlayson02,Finlayson06,Vicente-etal-PAMI18,vicentesingle}.

%Previous work model this color difference 
%\mtodo{chromatic of the light source, ECCV W with alex 2010 W, PAMI 2013} \mtodo{all critial is on the boundary, hue changed - need to infer, penumbera - need inpainting} \mtodo{Recently, make two physical obs in a fully superised setting Here we observe that the boundary patches can provide constraint. These apply for patches. one is ..., we address } 

By simply cropping shadow images into patches, we are posing the shadow removal as an unpaired image-to-image cross-domain mapping \cite{Yi2017DualGANUD,Choi2017StarGANUG,Liu2017UnsupervisedIT} that can be estimated via an adversarial framework. In particular, we seek a mapping function $G$ that takes as input a shadow-boundary patch $x$ from the set $\mB$, and outputs an image patch $\hat{x}$, such that a critic function $D$ cannot distinguish whether  $\hat{x}$ was drawn from the non-shadow set $\mN$ or  generated by $G$. Note that one potential solution here is to use Cycle-GAN or MaskShadow-GAN to estimate this transformation. However, the mapping functions learned by these methods are not able to remove shadows from patches in the full-shadow set $\mF$.  %2) The principle idea of MaskShadow-GAN is to copy-and-paste the shadow effects from a shadow image to a non-shadow template. Such mechanic would not be effective to self-cast shadows or shadows of aerial images since those type of shadows specifically associate to the objects in the scene. 3) There is no guarantee that  system would learn physically plausible shadow removal.
%Unpaired cross-domain mapping for shadow removal is extremely challenging. 

Training such an unpaired image-to-image mapping for shadow removal is challenging. The mapping is under-constrained and  training can collapse easily. \cite{gulrajani2017improved,liu2018two,Liu_2019_ICCV,Mescheder2018ICML,thanh2019improving,miyato2018spectral}. %MaskShadow-GAN takes advantage of the cycle-loss to stabilize and better approximate this shadow removal from data. 
Here, we propose to systematically constrain the shadow removal process by a physical model of shadow formation \cite{Shor08} and incorporate a number of physical properties of shadows into the framework. We show that these physics-based priors define a transformation closely modelling shadow removal. Driven by an adversarial signal, our framework effectively learns physically-plausible shadow removal without any direct supervision from paired data. Specifically, we constrain the shadow removal process to a shadow image decomposition model \cite{Le-etal-ICCV19} that extracts  a set of parameters and a matting layer from the shadow image. This set of shadow parameters is responsible for removing shadows on the umbra areas of the shadows via a linear function. Thus, once we estimate these shadow parameters from shadow boundary patches, we can use them to remove shadows for patches fully covered by the same shadow under the assumption that they share the same set of shadow parameters. %Compared to image-to-image translation, this mapping from shadow image to shadow parameters and shadow matte is simpler to train and easier to control in an adversarial training setting. 
%Importantly, the estimated shadow parameters computed from shadow boundary patches can be used to remove shadows for patches fully covered by the same shadow under the assumption that they share the same set of shadow parameters. %While such a shadow image decomposition model can be easily trained in a fully supervised setting as illustrated in \cite{Le-etal-ICCV19}, training it with just shadow masks is an extremely challenging task. 
%Note that in \cite{Le-etal-ICCV19}, all the shadow parameters and matting layers were pre-computed using the paired training images and the network was trained to simply regress those values. Here our model automatically estimates them through adversarial training. 
%\mtodo{in the shadow case, there exist a number of physic-based constaints / training practice, take home message, one-2-many mapping to fool D}
%We ensure proper shadow removal by multiple constraints and restrictions on each intermediate output component of the framework based on the physical properties of shadow. As 
Based on the physical properties of shadows, we apply the following constraints to the model:
\begin{itemize}
    \item We limit the search space of the shadow parameters and shadow matte to the appropriate value ranges that correspond to shadow removal.
    \item Our matting and smoothness losses  ensure that  shadow removal only happens in the shadow areas and transitions smoothly across shadow boundaries.
    \item Our boundary loss on the generated shadow-free image enforces  color similarity between the inner and outer areas alongside shadow boundaries.
\end{itemize}

%We use a smoothness and a matting loss to regularize the matte layer, and 
 %We propose a novel boundary loss to stabilize the training of the framework. After being trained on the patch level, our model can extract the shadow parameters using patches lying on the shadow boundary and use them to remove shadow for the patches fully covered by the same shadow. %We show that we can use the feedback from the discriminator to improve the quality of the generated images.
With these constraints and the adversarial signal, our method achieves  shadow removal results that are competitive with state-of-the-art  methods that were trained in a fully supervised manner with paired shadow and non-shadow images \cite{Le-etal-ICCV19,Wang_2018_CVPR,Qu_2017_CVPR}. We further compare our method to state-of-the-art methods on a novel and challenging video shadow removal dataset including static videos with various scenes and shadow conditions. This test exposes the weaknesses of data-driven methods trained on datasets lacking diversity. Our patch-based method seems to generalize better than other methods when evaluated on this video shadow removal test. Most importantly, we can easily fine-tune our pre-trained model on a single testing video to further improve  shadow removal results, showcasing this advantage of our training scheme. %This fine-tuning only requires shadow masks provided by a pre-trained shadow detector \cite{zhu18b}.  

In short, our contributions are:
\begin{itemize}
    \item We propose the use of an adversarial critic to train a shadow remover from unpaired shadow and non-shadow patches, providing an alternative solution to the paired data dependency issue.%, which eliminate the need of shadow-free data for training a shadow removal system.
    \item We propose a set of physics-based constraints that define a transformation closely modelling shadow removal, which enables shadow remover training with only an adversarial training signal.
    \item Our system trained without any shadow-free images has competitive results compared to fully-supervised state-of-the-art methods on the ISTD dataset. 
    \item We collect a novel video shadow removal dataset. Our shadow removal system can be fine-tuned for free to better remove shadows on testing videos. 
\end{itemize}

\section{Related Work}
\label{sec:related}
%\textbf{Shadow Removal.} 
Shadows are  physical phenomena. Early shadow removal works, without much training data, usually focused on studying different physical  shadow properties \cite{Finlayson06,Finlayson01,Finlayson02,Drew03recoveryof,Arbel2011ShadowRU,Fredembach2005HamiltonianPB,Liu2008TextureConsistentSR,Yang12}. Many works look for cues to remove shadows starting from shadow edges. Finlayson \etal \cite{Finlayson02} used shadow edges to estimate a scaling factor that differentiates  shadow areas from their non-shadow counterparts. Wu \& Tang \cite{Wu2005ABA} imposed a smoothness constraint alongside the shadow boundaries to handle  penumbra areas. Wu \etal\cite{Wu2012StrongSR} detected strong shadow-edges to remove shadows on the whole image. Shor \& Lischinki \cite{Shor08} defined an affine relationship between shadow and non-shadow pixels where they used the areas surrounding the shadow edges to estimate the parameters of such affine transforms. %Vicente \etal~\cite{Vicente-etal-PAMI18,vicentesingle} propose a color transfer procedure via histogram equalization for removing shadow. 
\par Shadow boundary effects  can also be modeled via image matting \cite{guoPami}. Wu \etal\cite{Wu07} estimated a matte layer representing the pixel-wise shadow probability to estimate a color transfer function to remove  shadows. Chuang \etal~\cite{Chuang2003} computed a shadow matte from video for shadow editing. They computed the lit and shadow images by finding min-max values at each pixel location throughout all frames of a video captured by a static camera. We use this technique to create a video dataset for testing shadow removal methods in Sec. \ref{sec:dataset}.

Current shadow removal methods \cite{Le-etal-ICCV19,Hu_2018_CVPR,Zhang:AAA2020,Ding2019ARGANAR,Wang_2018_CVPR} use deep-learning models  trained with full supervision on large-scale datasets~\cite{Wang_2018_CVPR,Qu_2017_CVPR}  of paired shadow and shadow-free images. Pairs are obtained by taking a photo with shadows, then removing the occluders from the scene to take the photo without shadows. Deshadow-Net \cite{Qu_2017_CVPR} extracted multi-context features to predict a matte layer that removes shadows. Some works use adversarial frameworks to train their shadow removal. In \cite{Wang_2018_CVPR} a unified adversarial framework predicted shadow masks and removed shadows. Similarly, Ding \etal \cite{Ding2019ARGANAR} used an adversarial signal to improve shadow removal in an iterative manner. Note that these methods use the shadow-free image as the main training signal while our method is trained only through an adversarial loss. In prior work \cite{Le-etal-ICCV19}  we constrained shadow removal by a physical model of shadow formation. We trained networks to extract shadow parameters and a matte layer to remove shadows. We adapt this model to patch-based shadow removal. Note that in \cite{Le-etal-ICCV19}, all  shadow parameters and matting layers were pre-computed using paired training images and the network was trained to simply regress those values, whereas our model automatically estimates them through adversarial training.  MaskShadow-GAN \cite{Hu_2018_CVPR} is the only deep-learning method that learns shadow removal from just unpaired training data.%, training on their dataset of unpaired shadow and shadow free images.
%\mtodo{transition,image-to-image}
%To provide an alternative solution 
%\textbf{Unpaired Image-to-Image Translation.} \mtodo{.}While traditional unpaired cross-domain mapping tasks \cite{CycleGAN2017,Yi2017DualGANUD} often aim to acquire visually plausible outputs, image-to-image cross-domain mapping for shadow removal is a many-to-one mapping where there is only one plausible shadow-free target. 

\section{Method}
We describe our patch-based shadow removal in Sec. \ref{sec:method_patch}. Our whole image pipeline for shadow removal  is described in Sec. \ref{sec:method_im}. For both image-level and patch-level shadow removal, we use  shadow matting \cite{Chuang2003,Porter1984,Smith1996,Wright} to express a shadow-free image $I^{\textrm{shadow-free}}$ by: 

\begin{equation}
    I^{\textrm{shadow-free}} =  I^{\textrm{relit}}\cdot \alpha+ I^{\textrm{shadow}} \cdot (1-\alpha )
    \label{eq:decom}
\end{equation}
with  $I^{\textrm{shadow}}$  the shadow image, $\alpha$ the matting layer, and $I^{\textrm{relit}}$ the relit image. The relit image contains shadow pixels relit to their non-shadow values,  computed via a linear function following a physical shadow formation model  \cite{Le-etal-ICCV19,Shor08}:
\begin{equation}
    I_i^{\textrm{relit}} = w \cdot I_i^{\textrm{shadow}} +b
    \label{eq:relit}
\end{equation}

The unknown factors in this shadow matting formula are the set of shadow parameters $(w,b)$ which define the linear function that removes the shadow effects in the umbra areas of the shadow, and the matte layer $\alpha$ that models the shadow effects on the shadow boundaries. We train a system of three networks to estimate these unknown factors via adversarial training. We use the annotated shadow segmentation masks for training. For testing, we obtain a segmentation mask for the image using the shadow detector proposed by Zhu \etal~ \cite{zhu18b}.

\subsection{Patch-based Shadow Removal}
\label{sec:method_patch}

\begin{figure}[t!]
	\centering
  \includegraphics[width=0.9\textwidth]{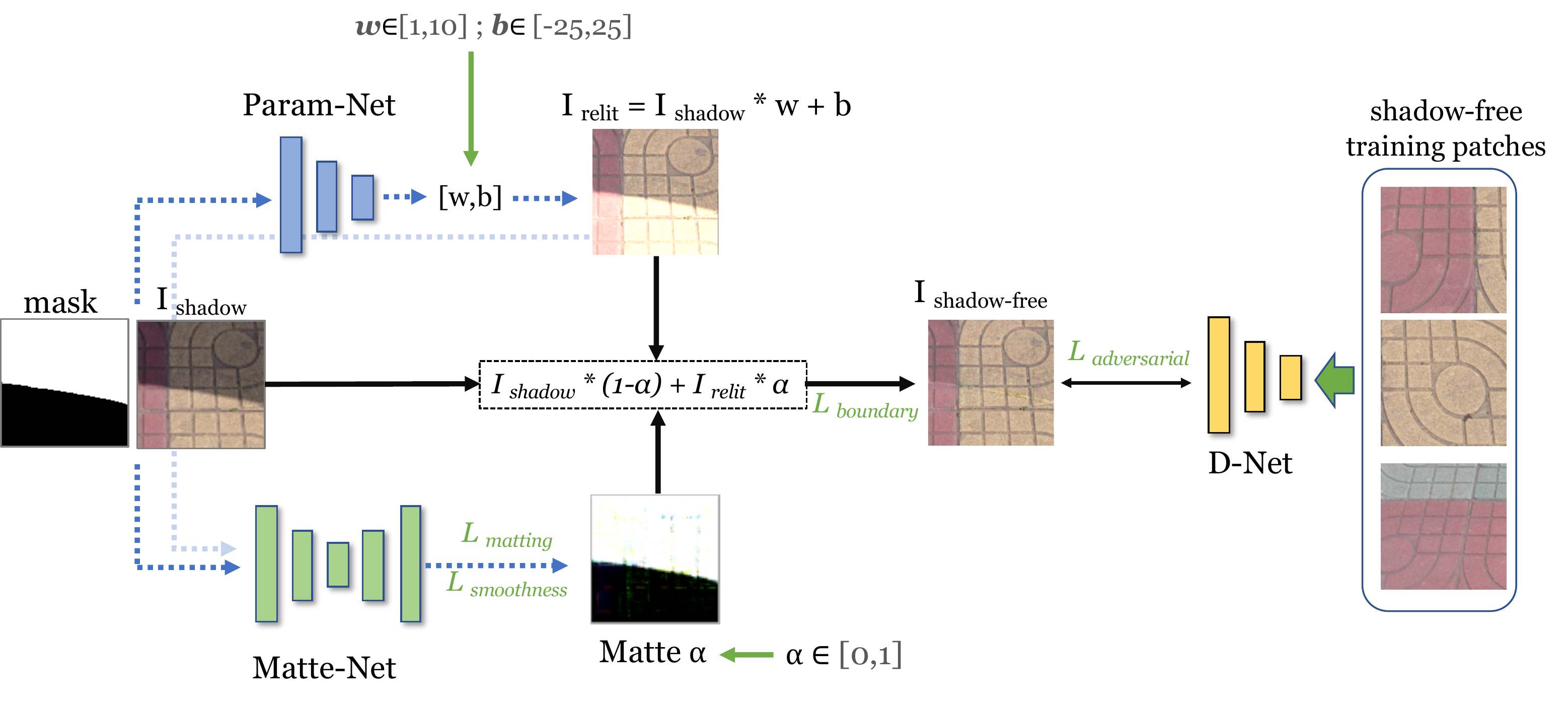}

  \caption{\textbf{Weakly-supervised shadow decomposition.} Our framework consists of three networks: Param-Net, Matte-Net, and D-Net. Param-Net and Matte-Net predict the shadow parameters $(w,b)$ and the matte layer $\alpha$ respectively to jointly remove the shadow. Param-Net takes as input the input image patch and its shadow mask to predict three sets of shadow parameters $(w,b)$ for the three color channels, which is used to obtain a relit image. The input image patch, shadow mask, and relit image are input into Matte-Net to predict a matte layer.  D-Net is the critic function distinguishing between the generated image patches and the real shadow-free patches. The only supervision signal is the set of shadow-free patches. The four losses guiding this training are the matting loss, smoothness loss, boundary loss, and adversarial loss.}
  \label{fig:model}
\end{figure}

Fig. \ref{fig:model} summarizes our framework to remove shadows from a single image patch, which consists of three networks: Param-Net, Matte-Net, and D-Net. Param-Net and Matte-Net predict the shadow parameters $(w,b)$ and the matte layer $\alpha$ respectively to jointly remove shadows. D-Net is the critic distinguishing between the generated image patches and the real shadow-free patches. With Param-Net and Matte-Net being the generators and D-Net being the discriminator, the three networks form an adversarial training framework where the main source of training signal is the set of shadow-free patches. 

In theory, as D-Net is trained to distinguish patches containing shadow boundaries from patches without any shadows, a natural solution to fool D-Net is to remove the shadows in the input shadow patches to make them indistinguishable from shadow-free patches. However, such an adversarial signal from D-Net alone often cannot guide the generators, (Param-Net and Matte-Net) to actually remove shadows. The parameter search space is very large and the mapping is extremely under-constrained. In practice, we observe that without any constraints, Param-Net tends to output consistently high values of $(w,b)$ as they would directly increase the overall brightness of the image patches, and Matte-Net tends to introduce artifacts similar to visual patterns frequently appearing in the non-shadow areas.
Thus, our main idea is to constrain this framework with physical shadow properties.  Constraining the output shadow parameters, shadow mattes, and combined shadow-free images, forces the networks to only transform the input images in a manner consistent with shadow removal. 

First, Param-Net estimates a scaling factor $w$ and an additive constant $b$, for each R,G,B color channel, to remove the shadow effects on the shadowed pixels in the umbra areas of the shadows via Eq. (\ref{eq:relit}). Here we hypothesize that the main component that explains the shadow effects is the scaling factor $w$. Accordingly, we bound its search space to the range $[1; s_{max}]$. The minimum value of $w=1$ ensures that the transformation always scales up the values of the shadowed pixels. We set the search space for $b$ to the range $[-c,c]$ where we choose a relatively small value of $c=25$ (the pixel intensity varies in the range [0,255]). Our intuition is to force the network to define the mapping mainly via the scaling factor $w$. We choose $s_{max}=10$. This upper bound of $w$ prevents the network from collapsing as $w$  increases. As we show in the ablation study, the network fails to learn a shadow removal without  proper search space limitation. %Note that we use three sets of shadow parameters $(w,b)$ for the three RGB color channels, similar to \cite{Le-etal-ICCV19}.

Matte-Net estimates a blending layer $\alpha$ that combines the shadow image patch and the relit image patch into a shadow-free image patch via Eq.\ref{eq:decom}. The value of a pixel $i$ in the output image patch, $I_i^{\textrm{output}}$, is computed as:  
\begin{equation}
I_i^{\textrm{output}} =  I_i^{\textrm{relit}}\cdot \alpha_i+ I_i^{\textrm{shadow}} \cdot (1-\alpha_i )
\end{equation}
We map the output of Matte-Net to [0,1] as $\alpha$ is being used as a matting layer and constrain the value of $\alpha_i$ as follows:
\begin{itemize}
    \item If \textit{i} indicates a non-shadow pixel, we enforce $\alpha_i=0$ so that the value of the output pixel $I_i^{\textrm{output}}$ equals its value in the input image $I_i^{\textrm{shadow}}$.
    \item If \textit{i} indicates a pixel in the umbra areas of the shadows, we enforce $\alpha_i=1$ so that the value of the output pixel $I_i^{\textrm{output}}$ equals its relit value $I_i^{\textrm{relit}}$.
    \item We do not control the value of $\alpha$ in the penumbra areas of the shadows and rely on the training of the network to estimate these values.
\end{itemize}

%These choices of values are obvious. A value of $\alpha_i=0$ results in pixels  
%Where  the umbra, non-shadow, or penumbra areas can roughly be specified using the shadow mask: 
%We define two areas alongside the shadow boundary, denoted as $\mM_{in}$ and $\mM_{out}$, 
\noindent where the umbra, non-shadow or penumbra areas can be roughly specified using the shadow mask. We define two areas alongside the shadow boundary, denoted as $\mM_{in}$ and $\mM_{out}$ - see Fig.\ref{fig:boundary}. $\mM_{out}$ is the area right outside the boundary, computed by subtracting the shadow mask, $\mM$, from its dilated version $\mM_{dilated}$. The inside area $\mM_{in}$ is computed similarly by subtracting an eroded shadow mask from the shadow mask. 
These two areas $\mM_{in}$ and $\mM_{out}$ roughly define a small area surrounding the shadow boundary, which can be considered as the penumbra area of the shadow. Then the above constraints are implemented as the matting loss $\mathcal{L}_{mat-\alpha} $ computed by the following formula for every pixel $i$:

\begin{equation}
\mathcal{L}_{mat-\alpha} = \sum_{i\in (\mM - \mM_{in})} |\alpha_i-1| +  \sum_{i\notin \mM_{dilated}} |\alpha_i|
\end{equation}

\def\subboxsize{0.31\textwidth}
\begin{figure}[t!]
	\centering
  \includegraphics[width=0.8\textwidth]{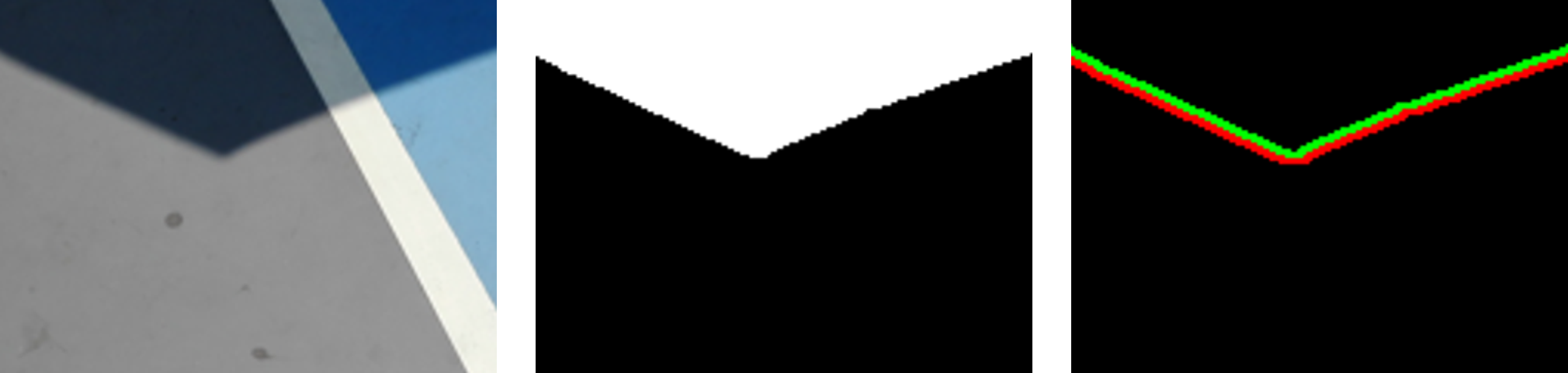}
    \makebox[\subboxsize]{Input Image}
    \makebox[\subboxsize]{Shadow Mask}
    \makebox[\subboxsize]{$\mM_{in}$ (green) \& $\mM_{out}$ (red)}

  \caption{\textbf{The penumbra area of the shadow.} We define two areas alongside the shadow boundary, denoted as $\mM_{in}$ (shown in green) and $\mM_{out}$ (shown in red). These two areas roughly define a small region surrounding the shadow boundary, which can be considered as the penumbra area of the shadow. }
  \label{fig:boundary}

\end{figure}

Moreover, since the shadow effects are assumed to vary smoothly across the shadow boundaries, we enforce an $L1$ smoothness loss on the spatial gradients of the matte layer, $\alpha$. This smoothness loss $\mathcal{L}_{sm} $ also prevents Matte-Net from producing undesired artifacts since it enforces local uniformity. This loss is:

\begin{equation}
\mathcal{L}_{sm-\alpha} =|\nabla{\alpha}| 
\end{equation}

Then, given a set of estimated parameters $(w,b)$ and a matte layer $\alpha$, we obtain an output image $I^{\textrm{output}}$ via the image decomposition formula (\ref{eq:decom}). We penalize the $L1$ difference between the average intensity of pixels lying right outside and inside the shadow boundary, which are the two areas $\mM_{in}$ and $\mM_{out}$. This shadow boundary loss $\mathcal{L}_{bd}$ is computed by:

\begin{equation}
\mathcal{L}_{bd} = \left|  \frac{\sum_{i\in \mM_{in}} I_i^{output}}{\sum_{i\in \mM_{in}} }  -  \frac{\sum_{i\in \mM_{out}}  I_i^{output}}{\sum_{i\in \mM_{out}}}\right|
\end{equation}

Last, we compute the adversarial loss with the feedback from D-Net: 
\begin{equation}
\mathcal{L}_{GAN} =\log (1-D(I^{output})) 
\end{equation}
where $D(\cdot)$ denotes the output of D-Net.

The final objective function to train Param-Net and Matte-Net is to minimize a weighted sum of the above losses:
\begin{equation}
\mathcal{L}_{final} = \lambda_{sm}\mathcal{L}_{sm-\alpha} + \lambda_{mat}\mathcal{L}_{mat-\alpha} + \lambda_{bd}\mathcal{L}_{bd}+ \lambda_{adv}\mathcal{L}_{GAN}
\end{equation}

All these losses are essential for training our networks, as shown in our ablation study in Sec. \ref{sec:ablation}. %Generally speaking, $\mL_{bd}$ keeps the shadow parameters in a limited range and stabilizes the training, $\mL_{sm-\alpha}$ constrains the matte layer to composite the relit image into the shadow image properly and reduces the artifacts, and $\mL_{mat-\alpha}$ ensures that the framework tries to remove the shadows rather than doing image inpainting. 
By using all the proposed losses together, our method is able to automatically extract a set of shadow parameters and an $\alpha$ layer from an input image patch. 
Fig. \ref{fig:decompose} visualizes the components extracted from our framework for two challenging input patches. %We visualize the relit images, computed from the estimated shadow parameters, and the matte layers. 
%The framework is able to accurately extract the shadow parameters and the matte layers even for challenging cases. 
In the first row, a dark shadow area is lit correctly to its non-shadow value. In the second row, the matte layer $\alpha$ is not affected by the dark material of the surface.
%In  supplementary material, we visualize the learned relit images and matte layers when each loss is removed.

\def\subfig{0.9\textwidth}
\def\subboxsize{0.15\textwidth}
\begin{figure*}[!h]
 \centering
 \includegraphics[width=\subfig]{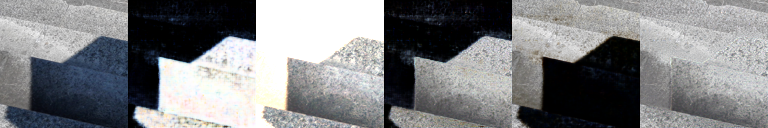}\\
 \includegraphics[width=\subfig]{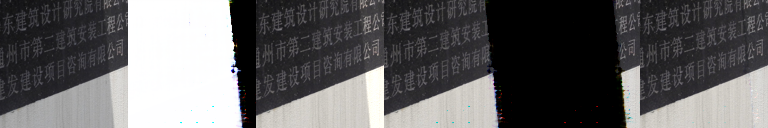}\\
    \makebox[\subboxsize]{$I^{sd}$}
    \makebox[\subboxsize]{$\alpha$}
    \makebox[\subboxsize]{$I^{relit}$}
    \makebox[\subboxsize]{$I^{relit}*\alpha$}
    \makebox[\subboxsize]{$I^{sd}*(1-\alpha)$ }
    \makebox[\subboxsize]{$I^{output}$}\\
     \caption{\textbf{Weakly-supervised shadow image decomposition.} With only shadow mask supervision, our method automatically learns to decompose the shadow effect in the input image patch $I^{sd}$ into a matte layer $\alpha$ and a relit image $I^{relit}$. The matte layer $\alpha$ combines $I^{sd}$  and $I^{relit}$ to obtain a shadow-free image patch $I^{output}$ via Eq. (\ref{eq:decom}).
    }
 
    \label{fig:decompose}

\end{figure*}

%\mtodo{figure to show effective of each loss}
\subsection{Image Shadow Removal using a patch-based model.}
\label{sec:method_im}

We estimate a set of shadow parameters and a matte layer for the input image to remove shadows via Eq. (\ref{eq:decom}). First, we obtain a shadow mask using the shadow detector of Zhu \etal~\cite{zhu18b}. We crop the input shadow image into overlapping patches. All patches containing the shadow boundaries  are then input into the three networks. We approximate the whole image shadow parameters from the patch shadow parameters, under the assumption that they share the same or very similar parameters. We simply compute the image shadow parameters as a linear combination of the patch shadow parameters. Similarly, we compute the values of each pixel in the matte layer by combining the overlapping matte patches. We set the matte layer pixels in the non-shadow area to $0$ and those in the umbra area to $1$. We observe that the classification scores obtained from the critic function D-Net correlate with the quality of the generated image patches. Thus, we normalize these scores to sum to 1 and use them as coefficients for the linear combinations that form the image shadow parameters and matte layer.

\section{Experiments}
\subsection{Network Architectures and Implementation Details.}
We use a VGG-19 architecture for Param-Net and a U-Net architecture for Matte-Net. D-Net is a simple 5-layer convolutional network. To map the outputs of the networks to a certain range, we use Tanh functions together with scaling and additive constants. We use stochastic gradient descent
with the Adam solver \cite{Adam} to train our model. The initial learning rate for Matte-Net and D-Net is 0.0002 and for Param-Net is 0.00002. All networks were trained from scratch. We experimentally set our training parameters ($\lambda_{bd},\; \lambda_{mat-\alpha},\; \lambda_{sm-\alpha},\; \lambda_{adv}$) to $(0.5,\; 100,\; 10,\; 0.5)$. We train our network with batch size 96 for 150 epochs. \footnote{All  code, trained models, and data are available at: \pink{\url{ https://www3.cs.stonybrook.edu/~cvl/projects/FSS2SR/index.html} }}

We use the ISTD dataset \cite{Wang_2018_CVPR} for training. Each original training image of size $640\times480$ is cropped into patches of size $128\times128$ with a step size of 32. This creates 311,220 image patches from 1,330 training shadow images. This training set includes 151,327 non-shadow patches, 147,312 shadow-boundary patches, and 12,581 full-shadow patches. 

\subsection{Shadow Removal Evaluation}

We first evaluate our method on the adjusted testing set of the ISTD dataset \cite{Wang_2018_CVPR,Le-etal-ICCV19}. % The testing set was adjusted \cite{Le-etal-ICCV19} to mitigate the color mismatch between the input shadow image and the ground-truth shadow-free image due to data acquisition setup. 
Following previous work \cite{Wang_2018_CVPR,guoPami,Qu_2017_CVPR,Le-etal-ICCV19}, we compute the root-mean-square-error (RMSE) in the LAB color space on the shadow area, non-shadow area, and the whole image, where all shadow removal results are re-sized to $256\times256$. Note that our method can  take any size image as input. We used the Zhu \etal~\cite{zhu18b} shadow detector, pre-trained on the SBU dataset and fine-tuned on the ISTD dataset, to obtain the shadow masks for our testing, as in \cite{Le-etal-ICCV19}. %We obtain the codes or the shadow removal results of these methods from the authors.

\setlength{\tabcolsep}{4pt}
\begin{table}[t]
\begin{center}
\caption{\textbf{Shadow removal results of our networks compared to state-of-the-art shadow removal methods on the adjusted ISTD testing set \cite{Le-etal-ICCV19,Wang_2018_CVPR}}. The metric is RMSE (the lower, the better). Best results are in bold.}
\label{table:headings}
\begin{tabular}{llccc}
\hline\noalign{\smallskip}
Methods   &Training Data                & Shadow& Non-Shadow& All  \\ 
\noalign{\smallskip}
\midrule
\noalign{\smallskip}
Input Image  & -             & 40.2  & 2.6 & 8.5\\ 
\midrule
Yang \etal~\cite{Yang12}   &   -           & 24.7  & 14.4 & 16.0\\ 
Guo \etal~\cite{guoPami}     &   Shd. Free + Shd. Mask  & 22.0  & 3.1 & 6.1\\ 
Gong \etal~\cite{Gong16}            &-& 13.3  & - & -\\ 
ST-CGAN \cite{Wang_2018_CVPR}  & Shd. Free + Shd. Mask  & 13.4  & 7.7 & 8.7\\ 
DeshadowNet \cite{Qu_2017_CVPR} & Shd. Free   & 15.9  & 6.0 & 7.6\\ 
MaskShadow-GAN \cite{hu_iccv2019mask}  &Shd. Free (Unpaired)        & 12.4  & 4.0 & 5.3\\ 
SP+M-Net \cite{Le-etal-ICCV19}& Shd. Free + Shd.Mask & \textbf{7.9}  &3.1 &\textbf{3.9}\\
\midrule
Ours & Shd. Mask & 9.7  &\textbf{3.0} &4.0\\
\hline
\end{tabular}
\end{center}
\end{table}
\setlength{\tabcolsep}{1.4pt}

In Table \ref{table:headings}, we compare our weakly-supervised methods with the recent state-of-the-art methods of Guo \etal~\cite{guoPami}, Gong \etal~\cite{Gong16}, Yang \etal~\cite{Yang12}, ST-CGAN \cite{Wang_2018_CVPR}, DeshadowNet \cite{Qu_2017_CVPR}, MaskShadow-GAN \cite{hu_iccv2019mask}, and SP+M-Net \cite{Le-etal-ICCV19}. The second column shows the training data of each method. All other deep-learning methods require paired shadow-free images as training signal except MaskShadow-GAN, which is trained on unpaired shadow and shadow-free images from the ISTD dataset.   
ST-CGAN and SP+M-Net also require the training shadow masks. Our method, trained without any shadow-free image, got 9.7 RMSE on the shadow areas, which is competitive with SP+M-Net. However, SP+M-Net requires full supervision.
\par Our method outperforms MaskShadow-GAN by 22\%, reducing the RMSE in the shadow area from 12.4 to 9.7 while also achieving lower RMSE on the non-shadow area. We outperform DeshadowNet and ST-CGAN, two methods that were trained with paired shadow and shadow-free images, reducing the RMSE by 38\% and 26\% respectively.

\def\subfig{0.95\textwidth}
\def\subboxsize{0.15\textwidth}
\begin{figure*}[t]
 \centering
\includegraphics[width=\subfig]{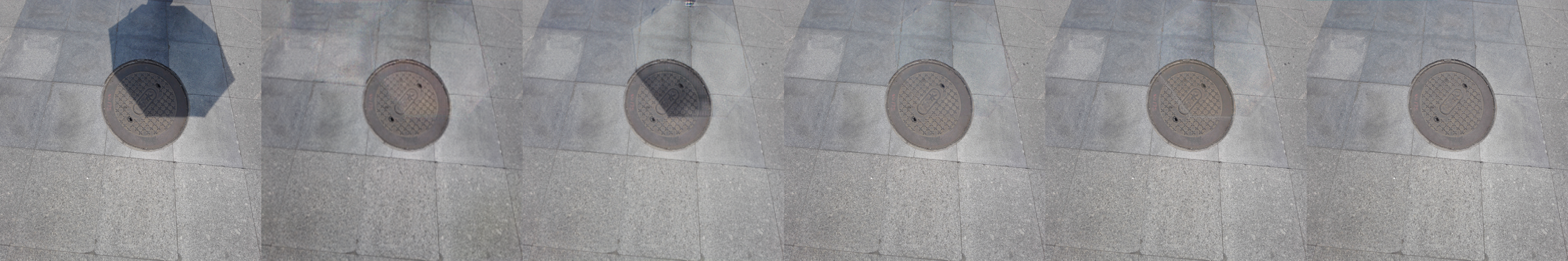}\\
\includegraphics[width=\subfig]{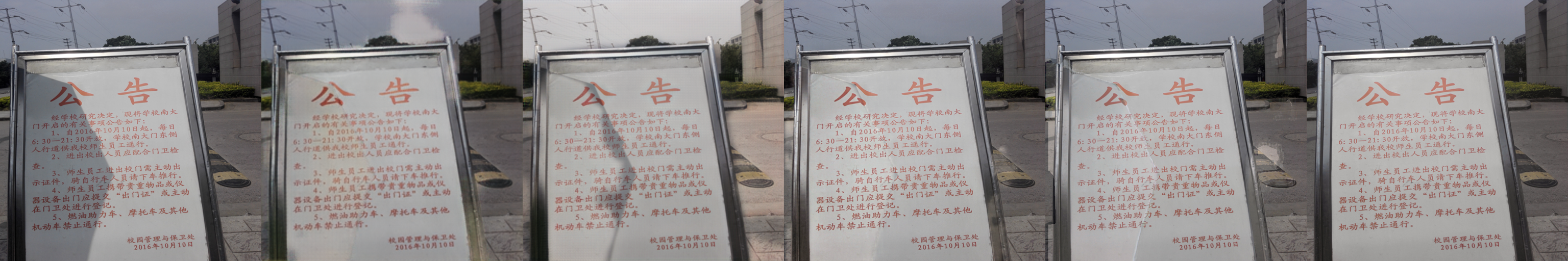}\\
\includegraphics[width=\subfig]{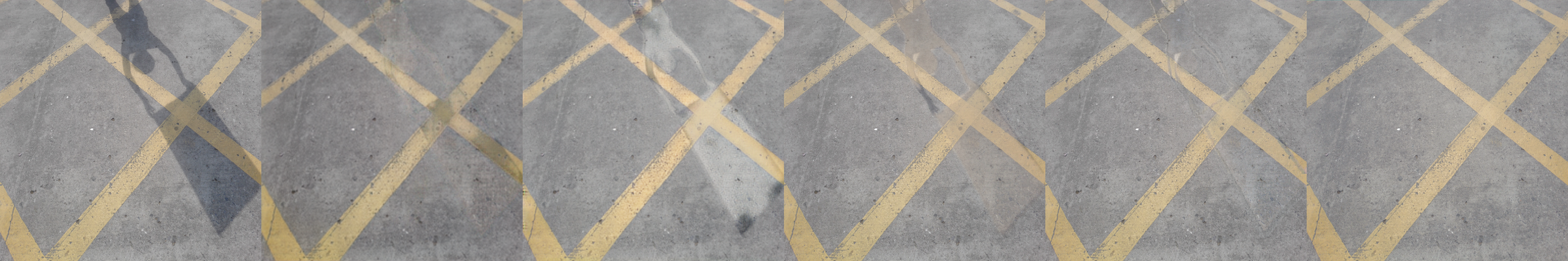}\\
\includegraphics[width=\subfig]{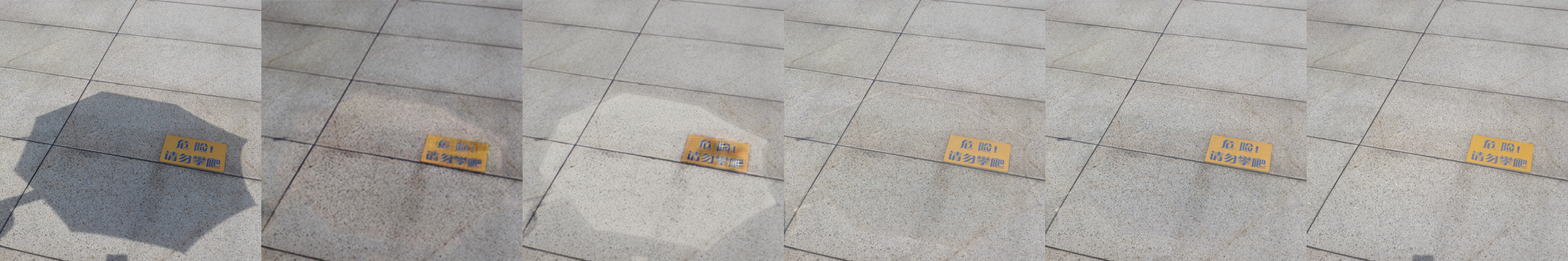}\\

    \makebox[\subboxsize]{Input}
   % \makebox[\subboxsize]{\cite{Qu_2017_CVPR}}
     \makebox[\subboxsize]{\cite{Wang_2018_CVPR}}
    \makebox[\subboxsize]{\cite{hu_iccv2019mask}}
    \makebox[\subboxsize]{\cite{Le-etal-ICCV19}}
    \makebox[\subboxsize]{Ours }
    \makebox[\subboxsize]{GT}\\
     \caption{{\bf Comparison of shadow removal on ISTD dataset.} Qualitative comparison between our method and the state-of-the-art methods: ST-CGAN \cite{Wang_2018_CVPR}, MaskShadow-GAN \cite{hu_iccv2019mask}, SP+M-Net\cite{Le-etal-ICCV19}. Our method, trained without any shadow-free images,  produces clean shadow-free images with very few artifacts.
    }
    \label{fig:main}
\end{figure*}
%Analysis on parameters learned, compared to linear regression. How does it correlate to RMSE

Fig. \ref{fig:main} compares qualitative shadow removal results from our method with other state-of-the-art methods on the ISTD dataset. Our method, trained with just an adversarial signal, produces clean shadow-free images with very few artifacts. On the other hand, ST-CGAN and MaskShadow-GAN tend to produce blurry images, introduce artifacts, and often relight the wrong image parts. Our method generates images which are visually similar to that of SP+M-Net. While SP+M-Net is less affected by the error in the shadow masks (shown in the 2nd row), our method generates images with more consistent colors between areas inside and outside the shadow boundaries (3rd and 4th rows). In all cases, our method preserves almost perfectly  the textures beneath the shadows (last row).% We provide additional results, analysis, and comparisons in the supplementary material.

\subsection{Ablation Studies}
\label{sec:ablation}

We conduct  ablation studies to better understand the effects of each proposed component in our framework. Starting from the original model with all the proposed features and losses, we train new  models removing the proposed components one at a time. %We also report the shadow removal performance with and without the critics score weighting described in Sec.\ref{sec:method_im}.
Table \ref{table:abl} summarizes these experiments. The first row shows the results of our model when we set the search space of the scaling factor $w$ to $[-10,10]$ and the search space of the additive constant $b$ to $[-255,255]$. In this case, the model collapses and consistently outputs uniformly dark images. Similarly, the model collapses when we omit the boundary loss $\mL_{bd}$. We observe that this loss is essential in stabilizing the training as it prevents the Param-Net from outputting consistently high values. 
\par The matting loss $\mL_{mat-\alpha}$ and  $\mL_{GAN}$ loss are critical for learning  proper shadow removal. We observe that without the matting loss $\mL_{mat-\alpha}$, the model behaves similarly to an image inpainting model where it tends to modify all parts of the images to fool the discriminator.
Last, dropping the smoothness loss $\mL_{sm}$ only results in a  slight drop in shadow removal performance, from 9.7 to 10.2  RMSE on the shadow areas. However, we observe more visible boundary artifacts on the output images without this loss. %We include more qualitative examples of these studies in the supplementary material.
\setlength{\tabcolsep}{4pt}
\begin{table}[t]
\begin{center}
\caption{\textbf{Ablation Studies.} We train our network without a certain loss or feature and report the shadow removal performances on the ISTD dataset \cite{Wang_2018_CVPR}. The metric is RMSE (the lower, the better). The table shows that all the proposed features in our model are essential in training for shadow removal.}
\label{table:abl}
\begin{tabular}{lccc}
\hline\noalign{\smallskip}
Methods       & Shadow& Non-Shadow& All  \\ 
\noalign{\smallskip}
\midrule
\noalign{\smallskip}
Input Image  & 40.2  & 2.6 & 8.5\\ 
\midrule
Ours w/o limiting search space  &47.5  &2.9&9.9\\
Ours w/o $\mL_{bd}$  &41.7  &3.9 &9.8\\ 
Ours w/o $\mL_{mat-\alpha}$  & 38.7 &3.1 &9.0\\
Ours w/o $\mL_{sm-\alpha}$  & 10.2  &2.8 &4.0\\
Ours w/o $\mL_{GAN}$ & 26.9  &2.9 &6.8\\
%Ours w/o D-Net based weighting & 10.5  &3.0 &4.2\\
\midrule
Ours  & 9.7  &3.0 &4.0\\
\hline
\end{tabular}
\end{center}
\end{table}
\setlength{\tabcolsep}{1.4pt}
\subsection{Video Shadow Removal}
\label{sec:dataset}

Video Shadow Removal is challenging for shadow removal methods. A video sequence has hundreds of frames with changing shadows. It is even harder for videos with a moving camera, moving objects, and illumination changes.

To better evaluate the performance of shadow removal methods in videos, we collected a set of 8 videos, each containing a static scene without visible moving objects. We cropped those videos to obtain clips with the only dominant motions caused by the shadows (either by  direct light motion or motion of the unseen occluders). As can be seen from the top row of Fig. \ref{fig:video_exp}, the dataset includes videos containing shadows cast by close-up occluders, far distance occluders, videos with simple-to-complex shadows, and shadows on various types of backgrounds and materials. Inspired by \cite{Chuang2003}, we propose a ``max-min'' technique to obtain a single pseudo shadow-free frame for each video: since the camera is static and there is no visible moving object in the frames, the changes in the video are caused by the moving shadows. We first obtain two images $V_{max}$ and $V_{min}$ by taking the maximum and minimum intensity values at each pixel location across the whole video. $V_{max}$ is then the image that contains the shadow-free values of pixels if they ever go out of the shadows. Similarly, their shadowed values, if they ever go into the shadows, are captured in $V_{min}$. Fig. \ref{fig:video_exp} shows these two images for a video named ``plant''. From these two images, we can trivially obtain a mask, namely moving-shadow $\mM$, marking the pixels appearing in both the shadow and non-shadow areas in the video: 
\begin{align}
	\mM_{i} = \left\{ 
	\begin{array}{ll}
		1 & \hspace{3ex}\textrm{if } V_{max,i}  > V_{min,i}  + \epsilon \\
		0 & \hspace{3ex}\textrm{otherwise},
	\end{array}
	\right. 
\end{align}
where we set a small threshold of $\epsilon=40$.  This method allows us to obtain pairs of shadow and non-shadow pixel values in the moving-shadow mask, $\mM$, for free.

\def\subfig{0.18\textwidth}
\def\subfigH{0.14\textwidth}
\def\subboxsize{0.18\textwidth}
\begin{figure*}[t]
 \centering
\includegraphics[width=\subfig,height=\subfigH]{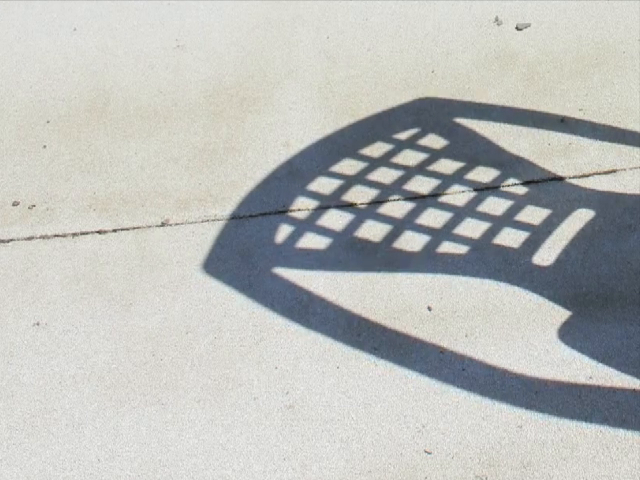}
\includegraphics[width=\subfig,height=\subfigH]{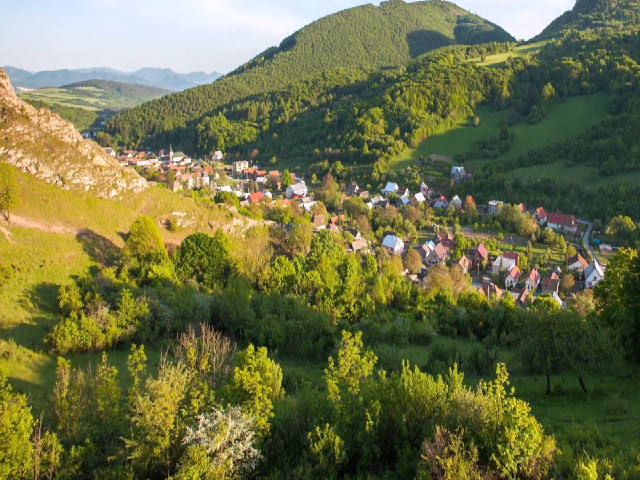}
\includegraphics[width=\subfig,height=\subfigH]{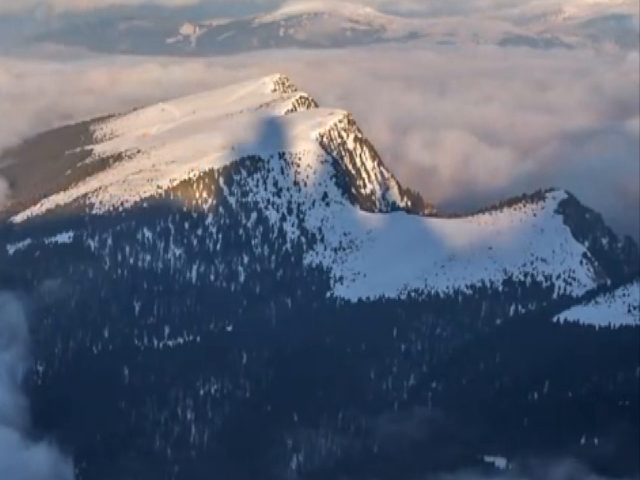}
\includegraphics[width=\subfig,height=\subfigH]{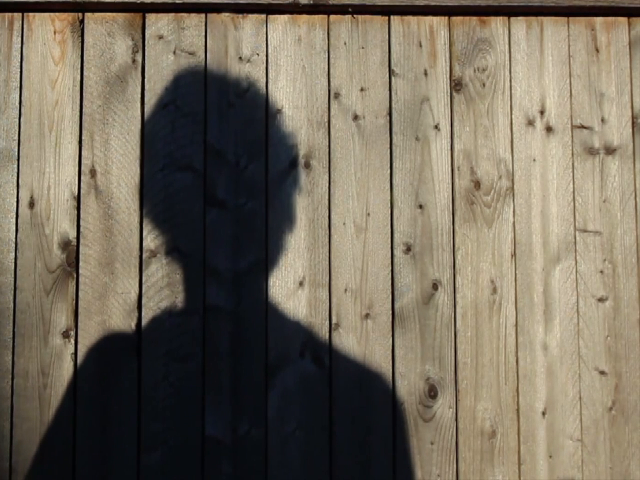}
\includegraphics[width=\subfig,height=\subfigH]{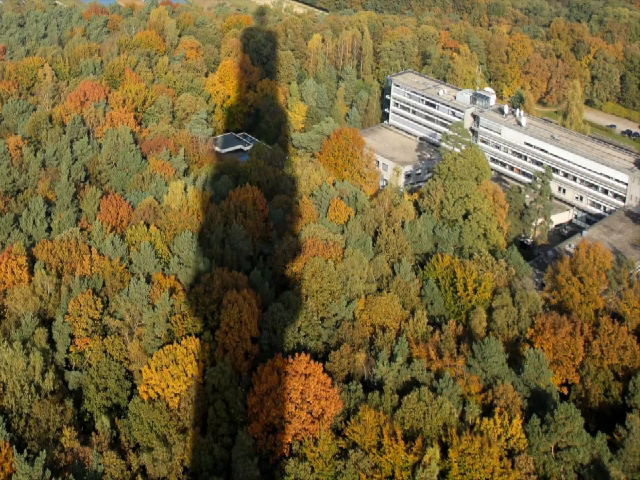}\\

  \makebox[\subboxsize]{}\\

\includegraphics[width=\subfig,height=\subfigH]{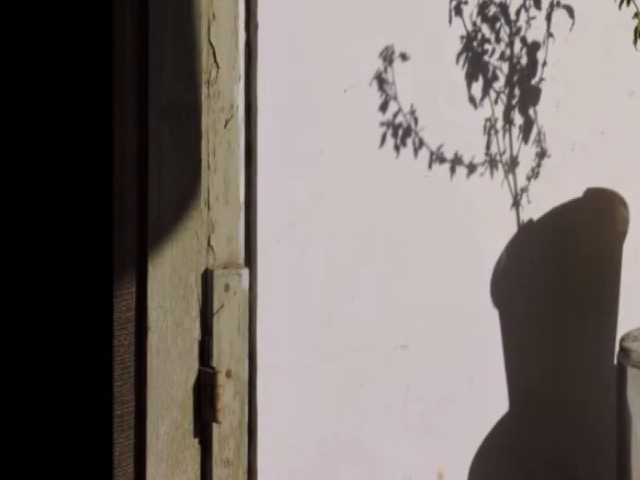}
\includegraphics[width=\subfig,height=\subfigH]{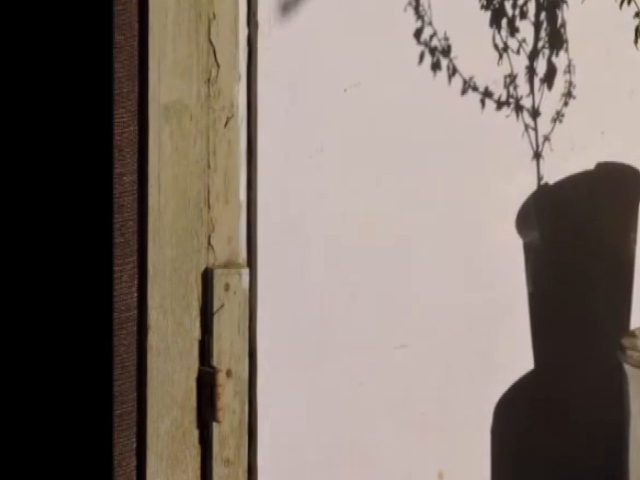}
\includegraphics[width=\subfig,height=\subfigH]{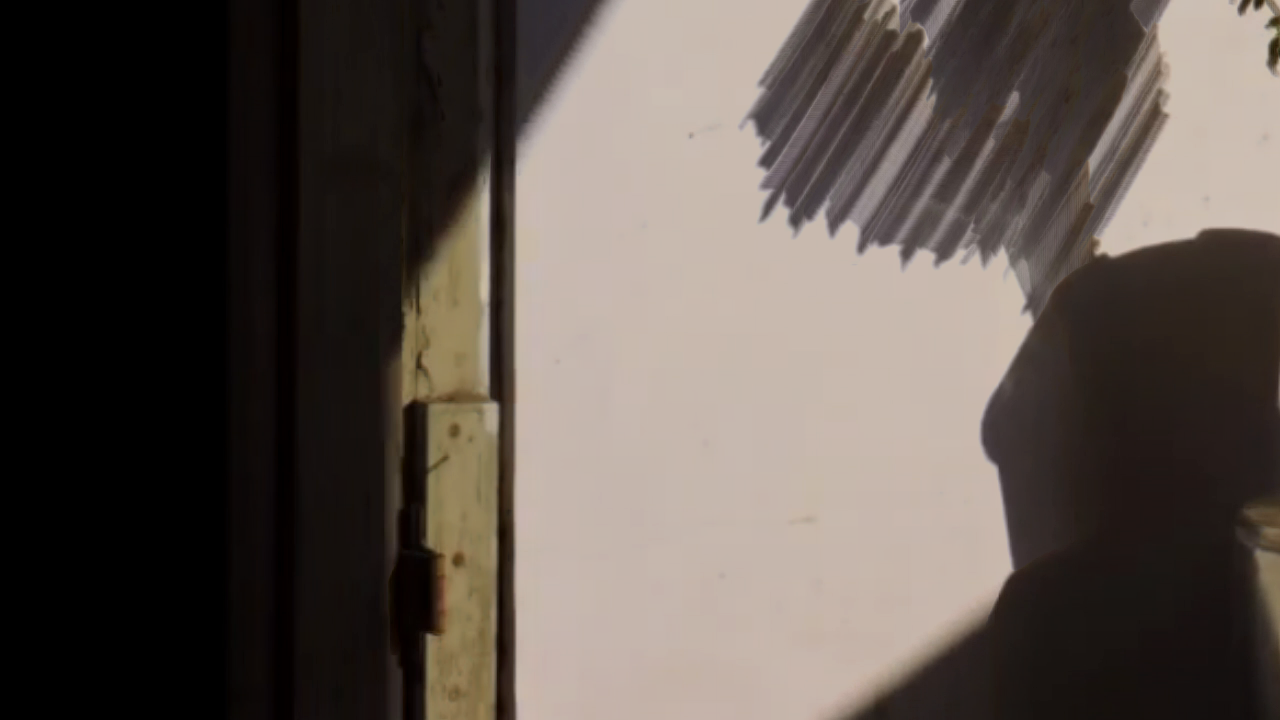}
\includegraphics[width=\subfig,height=\subfigH]{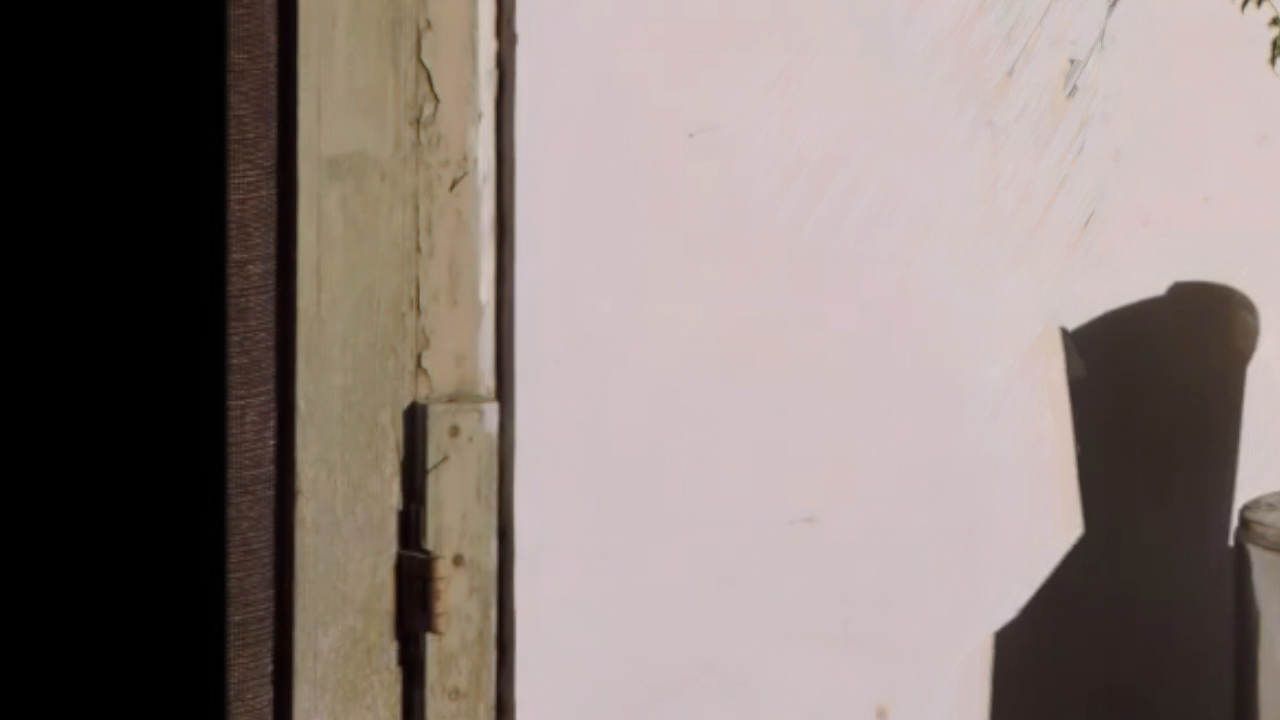}
\includegraphics[width=\subfig,height=\subfigH]{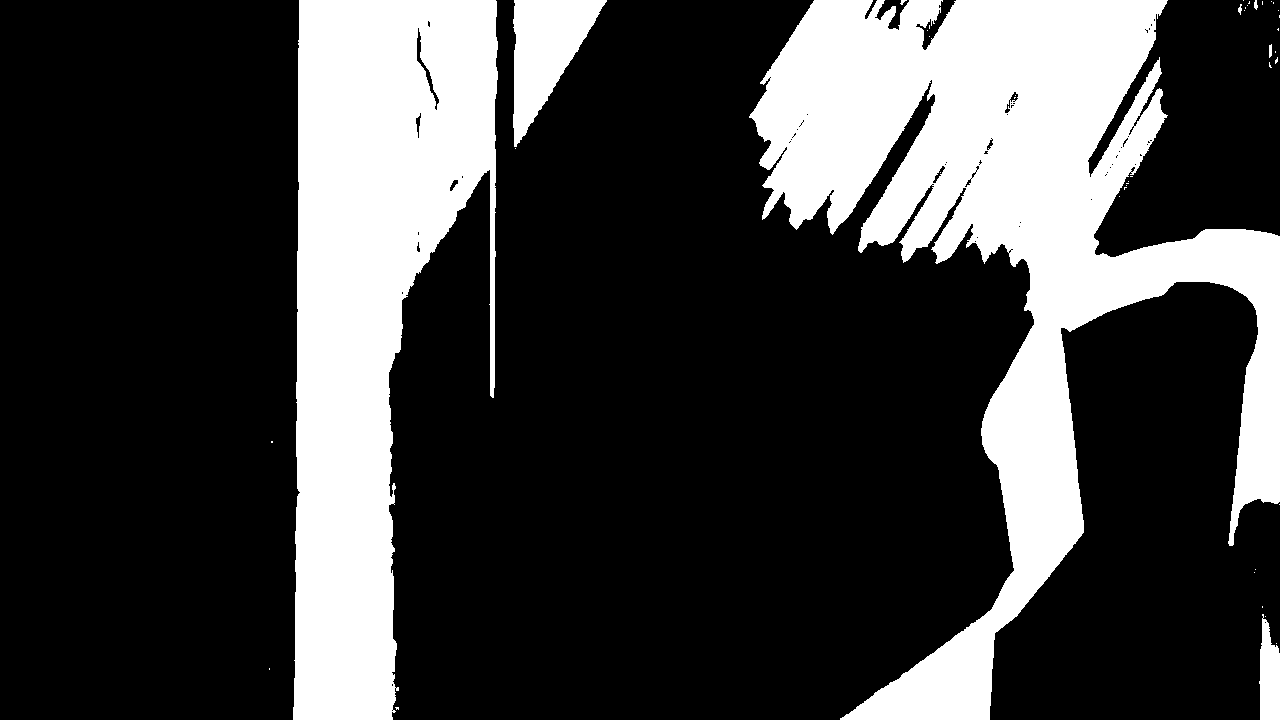}

     \makebox[\subboxsize]{Frame 0}
    \makebox[\subboxsize]{Frame 100}
    \makebox[\subboxsize]{$V_{min}$ }
    \makebox[\subboxsize]{$V_{max}$ }
    \makebox[\subboxsize]{Moving Shadow}\\
     \caption{{\bf Examples of Video Shadow Removal dataset.} The dataset consists of videos where both the scene and the visible objects remaining static. % The only source of motions in each video is the shadows. 
      The top row shows frames of different videos in our dataset. The second row visualizes our method to obtain the shadow-free frames for evaluating shadow removal.%: We first obtain $V_{max}$ and $V_{min}$ by taking the maximum and minimum values at each pixel location across the whole video. We use $V_{max}$ as a single pseudo shadow-free ground truth frame for the whole video. We measure the shadow removal performance only on the moving shadow mask areas to exclude the pixels that are covered in shadows for the whole video.  
    }
    \label{fig:video_exp}
\end{figure*}

To measure shadow removal performance, we input the frames of these videos into the shadow removal algorithm and measure the RMSE on the LAB color channel between the output frame and the image $V_{max}$ on the moving-shadow area $\mM$. We compute RMSE on each video and take their average to measure the shadow removal performance on the whole dataset. Table \ref{table:video} summarizes the performance of our methods compared to MaskShadow-GAN\cite{hu_iccv2019mask} and SP+M-Net\cite{Le-etal-ICCV19} on these videos. Our method outperforms SP+M-Net and MaskShadow-GAN, reducing the RMSE by 5\% and 11\% respectively. As our method only needs shadow segmentation masks for training, we use a pre-trained shadow detection model \cite{zhu18b} to obtain a set of shadow masks for each video. While these shadow mask sets are imperfect, fine-tuning our model using this free supervision results in  a 10\% error reduction, showing the advantage of our training scheme. Fig. \ref{fig:video} visualizes two example shadow removal results for different methods. We show a single input frame of each video. From left to right are the input frame, the shadow removal results of MaskShadow-GAN \cite{hu_iccv2019mask}, the results of SP+M-Net \cite{Le-etal-ICCV19}, the results of our model trained on the ISTD dataset, and the result of our model fine-tuned with each testing video for 1 epoch. The top row shows an example where all methods perform relatively well. Our method seems to have  better color balance between the relit pixels and the non-shadow pixels, although there is a visible boundary artifact due to imperfect shadow masks. After 1 epoch of fine-tuning, these artifacts are greatly suppressed. The bottom row shows a challenging case where all methods fail to remove shadows properly. %This example hints a weakness of our method as it relies on a relatively accurate shadow mask.

\setlength{\tabcolsep}{4pt}
\begin{table}[t]
\begin{center}
\caption{\textbf{Shadow removal results on our proposed Video Shadow Removal dataset}. The metric is RMSE (the lower, the better), compared to the pseudo shadow-free frame on the moving shadow mask. %Our method outperforms state-of-the-art methods on this challenging test. 
All methods were pre-trained on the ISTD dataset.  Ours+ denotes our model fine-tuned for one epoch on each video using the shadow masks generated by a shadow detector \cite{zhu18b} pre-trained on the SBU dataset\cite{Vicente-et-al-CVPR16}}
\label{table:video}
\begin{tabular}{lccccc}
\hline\noalign{\smallskip}
Methods &Input Frame & \cite{hu_iccv2019mask}& \cite{Le-etal-ICCV19}& Ours & Ours+ \\ 
\noalign{\smallskip}
\midrule
\noalign{\smallskip}
RMSE  & 32.9 & 23.5 & 22.2 & 20.9 & 18.0\\
\hline
\end{tabular}
\end{center}
\end{table}

\def\subfig{0.9\textwidth}
\def\subboxsize{0.19\textwidth}
\begin{figure*}[t]
 \centering
\includegraphics[width=\subfig,height=0.15\textwidth]{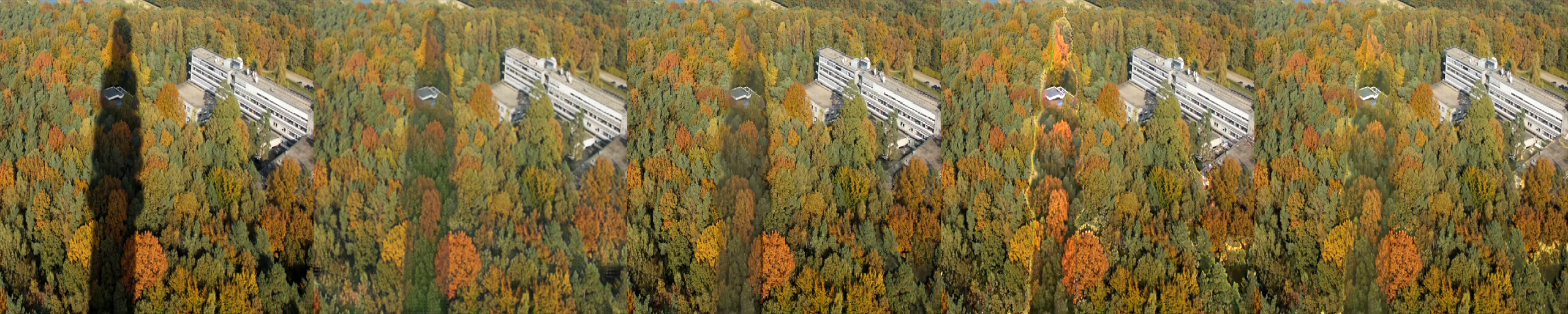}\\
\includegraphics[width=\subfig,height=0.15\textwidth]{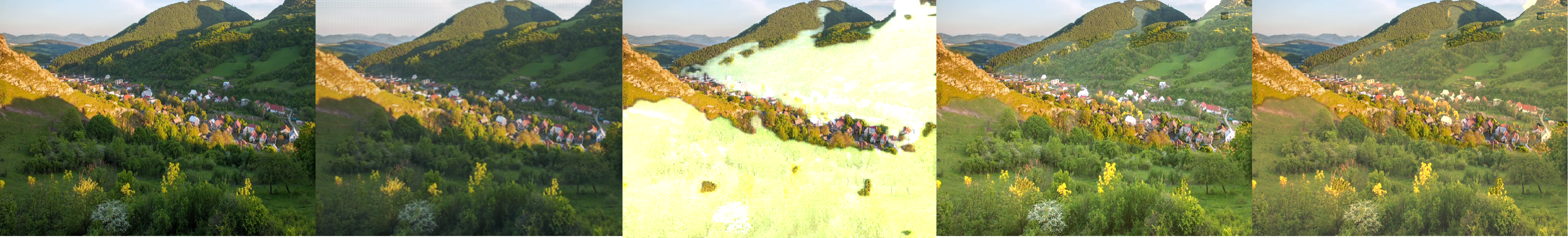}\\
    \makebox[\subboxsize]{Input Frame}
        \makebox[\subboxsize]{MaskShadow-GAN}
    \makebox[\subboxsize]{SP+M-Net}
    \makebox[\subboxsize]{Ours}
    \makebox[\subboxsize]{Ours+}\\
     \caption{{\bf Shadow Removal on Videos.}% From left to right are the input frame, the shadow removal results of MaskShadow-GAN \cite{hu_iccv2019mask}, the results of SP+M-Net \cite{Le-etal-ICCV19}, the results of our model trained on the ISTD dataset, and the result of our model fine-tuned with each testing video for 1 epoch. 
      We visualize the shadow removal results of different methods on two frames extracted from our video dataset. ``Ours+'' denotes the results of our model fine-tuned with each testing video for 1 epoch. 
     Top row shows an example where all methods perform relatively well. %Our method seems to have better color balance between the relit pixels and non-shadow pixels, albeit the visible boundary artifact due to the imperfect shadow masks. After 1 epoch of fine-tuning, these artifact are greatly suppressed. 
     Bottom row shows a challenging case where all methods fail to remove shadow properly.
    }
    \label{fig:video}
\end{figure*}
\setlength{\tabcolsep}{2pt}

\section{Conclusion}
We presented a novel patch-based deep-learning model to remove shadows from images. This method can be trained on patches cropped directly from the shadow images, using the shadow segmentation mask as the only supervision signal. This obviates the dependency on  paired training data and allows us to train this system on any kind of shadow image. The main contribution of this paper is a set of physics-based constrains that enable the training of this mapping. We have illustrated the effectiveness of our method on the standard ISTD dataset \cite{Wang_2018_CVPR} and on our novel Video Shadow Removal dataset. As shadow detection methods mature with the aid of recently proposed shadow detection datasets \cite{Wang2019InstanceSD,Hu2019RevisitingSD}, our method can be trained to remove shadows for a very low annotation cost.

\myheading{Acknowledgements.} 
This work was partially supported by the Partner University Fund, the SUNY2020 ITSC, and a gift from Adobe. Computational support provided by IACS and a GPU donation from NVIDIA.
We thank Kumara Kahatapitiya and 
Cristina Mata for assistance with the manuscript.

\clearpage
\bibliographystyle{splncs04}
\bibliography{shortstrings,longstrings,egbib}
\end{document}

%% file: definitions.tex
\def\mB{\mathcal{B}}

\def\mF{\mathcal{F}}

\def\mL{\mathcal{L}}
\def\mM{\mathcal{M}}
\def\mN{\mathcal{N}}

\def\1n{\mathbf{1}_n}
\def\0{\mathbf{0}}
\def\1{\mathbf{1}}

\definecolor{pink}{rgb}{0.9,0.5,0.5}
\definecolor{purple}{rgb}{0.5, 0.4, 0.8}   
\definecolor{gray}{rgb}{0.3, 0.3, 0.3}
\definecolor{mygreen}{rgb}{0.2, 0.6, 0.2}

\newcommand{\pink}[1]{\textcolor{pink}{#1}}

\definecolor{greena}{rgb}{0.4, 0.9, 0.1}

\definecolor{bluea}{rgb}{0, 0.4, 0.6}

\definecolor{reda}{rgb}{0.6, 0.2, 0.1}

\newcommand{\myheading}[1]{\vspace{1ex}\noindent \textbf{#1}}

\newcommand{\cm}[1]{}

\def\etal{\emph{et al}.}